%% file: naacl2021.tex
\pdfoutput=1

\documentclass[11pt]{article}

\usepackage[]{naacl2021}

\usepackage{times}
\usepackage{latexsym}

\usepackage[T1]{fontenc}

\usepackage[utf8]{inputenc}

\usepackage{microtype}

\usepackage{makecell}

\usepackage{microtype}

\usepackage{algorithm,algpseudocode,color,graphicx,multirow,paralist,pifont,subfigure,url,xspace}
\usepackage{amsfonts,amsmath}
\usepackage{booktabs}
\usepackage{setspace}
\definecolor{darkred}{rgb}{0.8,0.0,0.0}
\definecolor{darkblue}{rgb}{0.0,0.0,0.5}
\definecolor{darkgreen}{rgb}{0.0,0.5,0.0}
\newcommand{\textred}[1]{\textcolor{darkred}{#1}}
\newcommand{\textblue}[1]{\textcolor{darkblue}{#1}}
\newcommand{\textgreen}[1]{\textcolor{darkgreen}{#1}}
\usepackage{hyperref}
\usepackage{amssymb}
\usepackage{pifont}
\usepackage{color, colortbl}
\usepackage{soul}
\usepackage{color}
\usepackage{balance}

\definecolor{highlight}{gray}{0.85}

\title{\vspace*{-0.5in}
{\small \hfill Accepted to EMNLP 2021} \\
\vspace{0.35in}
Injecting Entity Types into Entity-Guided Text Generation}

\author{{\bf Xiangyu Dong$^{1*}$, Wenhao Yu$^{1*}$, Chenguang Zhu$^{2}$, Meng Jiang$^{1}$} \\
$^{1}$University of Notre Dame, Notre Dame, IN, \\
$^{2}$Microsoft Cognitive Services Research, Redmond, WA\\
\texttt{xdong2ps@gmail.com, wyu1@nd.edu} \\ \texttt{chezhu@microsoft.com, mjiang2@nd.edu}}

\begin{document}
\maketitle
\let\thefootnote\relax\footnotetext{* The first two authors have equal contributions.}
\let\thefootnote\relax\footnotetext{$\S$ Our code and datasets are available at \url{https://github.com/DM2-ND/InjType}. }

\begin{abstract}
\input{0abstract}
\end{abstract}

\section{Introduction}
\label{sec:introduction}
\input{1introduction}

\section{Related Work}
\label{sec:related}
\input{2related}

\section{Proposed Method: \textit{InjType}}
\input{3method}

\section{Experiments}
\label{sec:Experiments}
\input{4experiments}

\section{Conclusions}
\label{sec:conclusions}
\input{5conclusions}

\section*{Acknowledgements}
This work is supported by National Science Foundation IIS-1849816 and CCF-1901059.

\balance
\bibliography{reference}
\bibliographystyle{acl_natbib}

\clearpage
\input{6appendix}


\end{document}

%% file: 0abstract.tex
Recent successes in deep generative modeling have led to significant advances in natural language generation (NLG). Incorporating entities into neural generation models has demonstrated great improvements by assisting to infer the summary topic and to generate coherent content. To enhance the role of entity in NLG, in this paper, we aim to model the entity type in the \textit{decoding} phase to generate contextual words accurately. We develop a novel NLG model to produce a target sequence based on a given list of entities. Our model has a multi-step decoder that \emph{injects} the entity types into the process of entity mention generation. 
Experiments on two public news datasets demonstrate type injection performs better than existing type embedding concatenation baselines. 


%% file: 1introduction.tex

Entity, as an important element of natural language, plays the key role of making the text coherent~\cite{grosz1995centering}. Recently, modeling entities into NLG methods has demonstrated great improvements by assisting to infer the summary topic~\cite{amplayo2018entity} or to generate coherent content~\cite{ji2017dynamic,clark2018neural}. To enhance the representation of entity, entity type is often used in existing work -- represented as a separate embedding and concatenated with the embedding of entity mention (i.e., surface name) in the encoding/decoding phase~\cite{zhao2019personalized,puduppully2019data,yu2018typesql,chan2019stick}. Although
the concatenation performed better than using the entity mention embedding only, the relationship between entity mention and entity type was not reflected, making the signal from entity type undermined in the NLG.

\begin{table}[t]
\caption{An example of generating news with a list of names of entities and their types. How to use entity type information in NLG models is an open question.}
\vspace{-0.1in}
\centering
\scalebox{0.90}{\begin{tabular}{p{7.9cm}}
\toprule
    \textbf{Input:} [\textsc{Country}:US$_1$, \textsc{Person}:Dick\_Cheney$_2$, \textsc{Country}:Afghanistan$_3$, \textsc{Weekday}:Monday$_4$, \textsc{Country}:Afghan$_5$, \textsc{Person}:Hamid\_Karzai$_6$, \textsc{Organization}:NATO$_7$, \textsc{City}:Bucharest$_8$]  \\
\midrule
    \textbf{Target:} ``US$_1$ vice president Dick\_Cheney$_2$ made a surprise visit to Afghanistan$_3$ on Monday$_4$ for talks with Afghan$_5$ president Hamid\_Karzai$_6$, ahead of the NATO$_7$ summit early next month in Bucharest$_8$.'' \\
\bottomrule
\end{tabular}}
\vspace{-0.1in}
\label{tab:example}
\end{table}

\begin{figure*}[t]
{\includegraphics[width=1.0\textwidth]{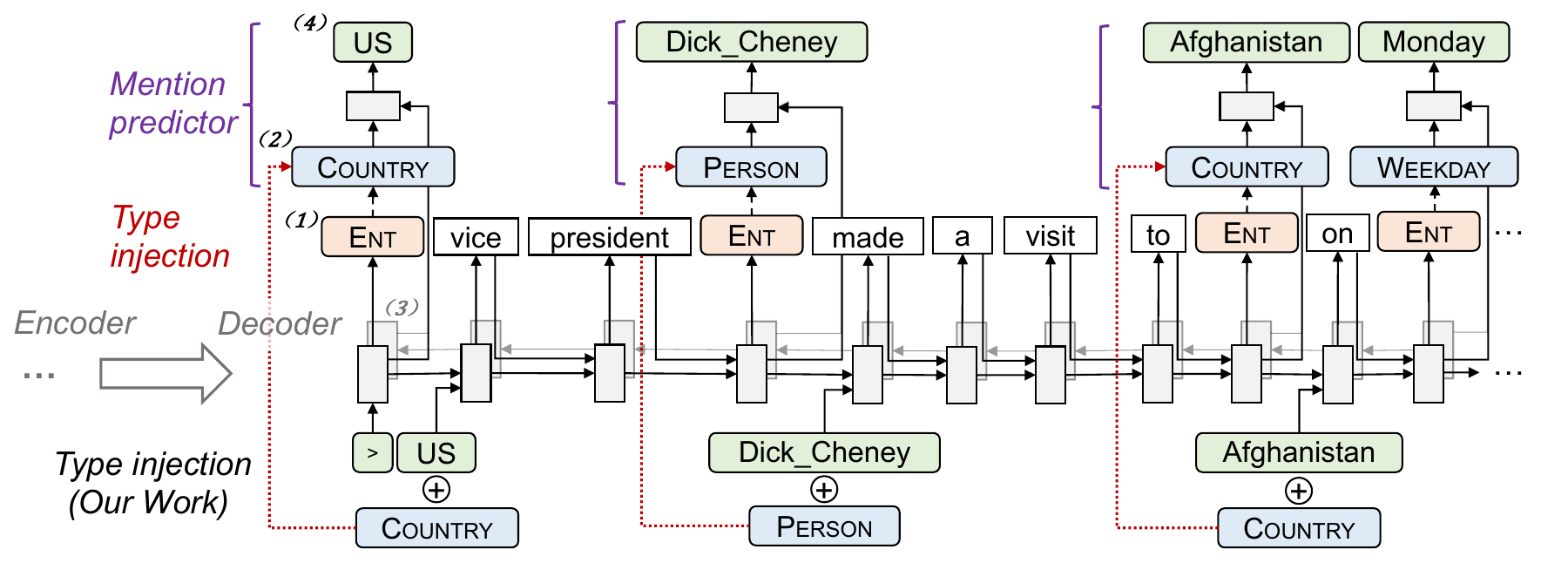}
\label{fig:framework_injection}}
\vspace{-0.3in}
\caption{
The \textbf{decoding} process of InjType has four steps: (S1) predicting the {<Ent>} token (i.e., entity indicator); (S2) injecting the entity types; (S3) combining an entity type enhanced NLU with backward information of target sequence; (S4) predicting the entity mention using the type embedding and hidden state by a mention predictor.}
\vspace{-0.05in}
\label{fig:frameworks}
\end{figure*}

To address the above issue, our idea is to model the entity type carefully in the decoding phase to generate contextual words accurately. In this work, we focus on developing a novel NLG model to produce a target sequence based on a given list of entities. 
Compared to the number of words in the target sequence, the number of given entities is much smaller.
Since the source information is extremely insufficient, it is difficult to generate precise contextual words describing the relationship between or event involving multiple entities such as person, organization, and location. 
Besides, since input entities are important prompts about the content in the target sequence~\cite{yao2019plan}, the quality of generated sequence depends significantly on whether the input entities are logically connected and expressed in the output. However, existing generation models may stop halfway and fail to generate words for the expected entities, leading to serious incompleteness~\cite{feng2018topic}.

In this paper, we propose a novel method of utilizing the type information in NLG, called \emph{InjType}. It keeps the same encoder as Seq2Seq models~\cite{sutskever2014sequence}.
During decoding, it first predicts the probability that each token is a contextual word in the vocabulary or an entity from a given list. If the token is an entity, the model will directly inject the embedding of the entity type into the process of generating the entity mention by using a mention predictor to predict the entity mention based on the type embedding and current decoding hidden state. The type injection maximizes the likelihood of generating an entity indicator rather than the likelihood of sparse entity mentions. The hidden state is jointly optimized by predicting the role of token and predicting the entity mention so the entity's information is effectively embedded into the hidden states. 

Experiments on two public news datasets \textsc{Gigawords} and \textsc{NYT} demonstrate that InjType can generate more precise contextual words than the existing concatenation-based models.

%% file: 2related.tex
\noindent\textbf{Entity-related Text Generation.}
Entities in a natural language carry useful contextual information~\cite{nenkova2008entity} and therefore play
an important role in different NLG tasks such as summarization~\cite{sharma2019entity,amplayo2018entity}, concept generation~\cite{zeng2021enhancing}, table description~\cite{puduppully2019data} and news generation~\cite{yu2021sentence}. In summarization, entity mentions have been used to extract non-adjacent yet coherent
sentences, link to existing knowledge bases, and infer the summary topic~\cite{sharma2019entity,amplayo2018entity}. In table description, entity mentions have been used to achieve discourse coherence~\cite{puduppully2019data}. Our task is relevant to \cite{chan2019stick} that generates product description from a list of product entities. Different from above work, we aim to leverage entity class into the decoding phase for better predicting entities and contextual words.

\vspace{0.05in}
\noindent\textbf{Words-to-text Generation.}
It is also referred as constraint text generation~\cite{zhang2020pointer,qin2019conversing}. 
Generating text from topic words and keywords is a popular task in NLG. It not only has plenty of practical applications, e.g., benefiting intelligent education by assisting in essay writing~\cite{feng2018topic,yang2019enhancing} and automated journalism by helping news generation~\cite{zheng2017automatic,zhang2020pointer}, but also serves as an ideal test bed for controllable text generation~\cite{wang2018sentigan,yao2019plan}. The main challenge of words-to-text lies in that the source information is extremely insufficient compared to the target output, leading to poor topic consistency in generated text~\cite{yu2020survey}.

%% file: 3method.tex
In this section, we first give the task definition, then introduce our proposed type injection method. \textbf{\underline{We note that}} \textit{InjTyp} users the same encoder as in Seq2Seq models~\cite{sutskever2014sequence}, i.e., a bi-directional GRU. So, in Figure \ref{fig:frameworks} and the following sections, we only describe the decoding process.

\vspace{0.05in}
\noindent\textbf{Task Definition}
Given a list of entities $X=(x_1, \ldots, x_n)$, where $x_i=(x^M_i \in \mathcal{M}, x^T_i \in \mathcal{T})$ consists of the mention and type of the $i$-th entity, where $\mathcal{M}$ is the set of entity mentions and $\mathcal{T}$ is the set of entity type. The expected output sequence is $y=(y_1, \dots, y_m)$ containing all the entity mentions.
We denote the vocabulary of contextual words by $\mathcal{V}$. So $y_j \in \mathcal{M} \cup \mathcal{V}$, $j \in \{1, \ldots, m\}$. The task is to learn a predictive function $f: X \rightarrow Y$, mapping a list of entities to a target sequence.

\subsection{Entity Indicator Predictor}

At each step, the decoder predicts either an entity indicator or a contextual word. An entity indicator, denoted as <Ent>, indicates that the current decoding step should generate an entity in the output sequence. If the input has $n$ entities, there will be $n$ entity indicator <Ent> generated in the output sequence. So the first-step output sequence is:
\vspace{-0.05in}
\begin{align}
    \mbox{block}_1, \mbox{{Ent}}_1, \mbox{block}_2, \ldots, \mbox{{Ent}}_n, \mbox{block}_{n+1}. 
    \nonumber
\end{align}

Each block has one or multiple contextual words, and it ends with an entity indicator (<Ent>). In each block, the generation process is the same as the auto-regressive decoding process. When the auto-regressive decoder generates an <Ent>, the generation process of the current block ends. 
When the decoder generated the ($n+1$)-th entity indicator <Ent>, the entire generation terminates. 

Suppose the ground truth of entity indicator output $y^{\prime\prime}$ is the target sequence with entity mentions replaced with entity indicators <Ent>. Now, the loss function with entity indicator is defined as:
\vspace{-0.05in}
\begin{equation}
    \mathcal{L}_{\text{Ent}} = - \sum^m_{t=1} \log \left( p(y^{\prime\prime}_t \in \{ \mbox{{Ent}} \} \cup \mathcal{V}|y^{\prime\prime}_{< t}, X) \right). \nonumber
\label{kbnll}
\end{equation}

\subsection{Mention Predictor}

Since each block's generation is ended with the entity indicator token (<Ent>), the representations of the last hidden states in different blocks are assimilated, which may lose contextual information in previous generated tokens. In order to let the decoding hidden states carry rich contextual information and better predict the entity mention, we present a novel mention predictor by \textit{injecting} the entity type embedding into current hidden state, and feed the combined embedding into a mention classifier. In $i$-th block, the predicted entity mention is $x^{M}_{i^\prime} = \mathrm{softmax}(\textbf{W}_m \cdot [\textbf{s}_t \oplus \textbf{x}^T_i])$, where $\textbf{s}_t$ is the hidden state of the $t$-{th} token in the generated text, and $\textbf{x}^T_i$ is the $i$-th entity type embedding.
In this way, the last hidden state in each block not only has to be classified as an entity indicator (<Ent>), but also carries both entity type and entity mention information in order to make precise generation. The classification loss $\mathcal{L}_{MP}$ can be written as:
\vspace{-0.1in}
\begin{equation}
    \mathcal{L}_{\text{MP}} = -\sum_{i=1}^n ~x^{M}_{i} \cdot \log(x^{M}_{i^\prime}), \nonumber
\end{equation}

\subsection{Entity type-Enhanced NLU} 

Inspired by the UniLM~\cite{dong2019unified}, we let our decoder complete a \textit{type enhanced NLU task} along with its original generation task. We borrow the decoder from the NLG task to conduct an NLU task on the ground truth articles during training. The entity type enhanced NLU task asks the decoder to predict entity mentions corresponding to the types in the ground truth based on context words. If the decoder is able to correctly predict the entity mention given contextual information, it should be capable of generating good context words that can help predict entity mentions as well. Since the decoder used for generation is naturally one-way (left-to-right), in order to complete the NLU task more reasonably, we train a GRU module in a reversed direction, represented as $\overleftarrow{\mathrm{GRU}}$. We reuse the original NLG decoder without attention, denoted by $\overrightarrow{\mathrm{GRU}^\prime}$ for the NLU task. This module generates the prediction as follows:
\begin{align}
    \textbf{s}^\prime_t = [ \overrightarrow{\mathrm{GRU}^\prime}(y^{\prime\prime}_t) \oplus \overleftarrow{\mathrm{GRU}}(y^{\prime\prime}_t)], \nonumber
\end{align}
where ${\textbf{s}^\prime_t}$ is the concatenated hidden state of the original hidden state $\textbf{s}_t$ and new hidden state derived from added GRU module. The entity mention is then predicted as $x^{M}_{i^{\prime\prime}} = \mathrm{softmax} (\textbf{W}_r \cdot \textbf{s}^{\prime}_t)$.
So the NLU loss is only calculated at the entity positions:
\vspace{-0.1in}
\begin{equation}
    \mathcal{L}_{\text{NLU}} = -\sum_{i=1}^n ~x^{M}_{i} \cdot \log(x^{M}_{i^{\prime\prime}}). \nonumber
\label{kbnll}
\end{equation}

\subsubsection{Joint Optimization}

InjType jointly optimizes the following loss:
\begin{equation} \label{eq:loss}
    \mathcal{L} = \mathcal{L}_{Ent} + \lambda_1 \cdot \mathcal{L}_{MP} + \lambda_2 \cdot \mathcal{L}_{NLU},
\end{equation}
where $\lambda_1$ and $\lambda_2$ are hyperparameters to control the importance of different tasks.

%% file: 4experiments.tex
\begin{table}[hb]
\vspace{-0.15in}
\caption{Statistics of two datasets. Additional information is in Section \ref{sec:data} and Table \ref{tab:datasets} in Appendix.}
\vspace{-0.1in}
\centering
\scalebox{0.88}{\begin{tabular}{l|c|c}
\toprule
& \textsc{gigawords-6k} & \textsc{NYT-8k} \\
\midrule
\#Articles & 7,903 & 8,371\\
\#Mentions $|\mathcal{M}|$ & 14,645 & 15,300\\
\#Types $|\mathcal{T}|$ & 14 & 14 \\
\#Words $|\mathcal{V}|$ & 30,121 & 38,802 \\
Len. Input/Output & 14.0 / 86.1 & 10.6 / 78.6 \\
\bottomrule
\end{tabular}}
\label{tab:datasets}
\vspace{-0.2in}
\end{table}

\begin{table*}[t]
\begin{center}
\caption{Our \textit{InjType} can outperform various baseline models enhanced by type embedding concatenation. 
}
\vspace{-0.1in}
\setlength{\tabcolsep}{2.5mm}{\scalebox{0.90}{{\begin{tabular}{l|rcc|rcc}
\toprule
{\multirow{2}*{Methods}}
& \multicolumn{3}{c|}{{\textsc{Gigawords}}}  & \multicolumn{3}{c}{{\textsc{NYT}}} \\
& \multicolumn{1}{c}{{ROUGE-2}} & \multicolumn{1}{c}{{ROUGE-L}} & 
\multicolumn{1}{c|}{{BLEU-4}} & \multicolumn{1}{c}{{ROUGE-2}} & \multicolumn{1}{c}{{ROUGE-L}} &
\multicolumn{1}{c}{{BLEU-4}} \\
\midrule
Seq2Seq & 8.83$\pm$0.15 & 31.43$\pm$0.13 & 12.21$\pm$0.30 & 8.83$\pm$0.15 & 31.43$\pm$0.13 & 12.21$\pm$0.30 \\
SeqAttn & 9.10$\pm$0.13 & 36.62$\pm$0.11 & 16.17$\pm$0.28 & 5.95$\pm$0.15 & 29.67$\pm$0.06 & 11.86$\pm$0.15 \\
CopyNet & 9.44$\pm$0.11 & 36.96$\pm$0.10 & 16.40$\pm$0.24 & 6.25$\pm$0.14 & 30.58$\pm$0.09 & 11.96$\pm$0.14 \\
GPT-2 & 9.04$\pm$0.20 & 31.30$\pm$0.16 & 15.66$\pm$0.40 & 5.86$\pm$0.20 & 24.19$\pm$0.14 & 10.89$\pm$0.22 \\
UniLM & 11.77$\pm$0.18 & 36.54$\pm$0.15 & 17.66$\pm$0.35 & 7.47$\pm$0.15 & 30.66$\pm$0.13 & 12.90$\pm$0.20 \\
\midrule
\textbf{\textit{InjType}} & \textbf{13.37$\pm$0.12} & \textbf{41.16$\pm$0.31} & \textbf{18.55$\pm$0.09} & \textbf{8.55$\pm$0.09} & \textbf{31.53$\pm$0.17}  & \textbf{13.14$\pm$0.03}  \\
~$\vdash$ w/o MP & 9.39$\pm$0.16 & 38.34$\pm$0.10 & 16.36$\pm$0.25 & 6.52$\pm$0.09 & 30.10$\pm$0.08 & 12.19$\pm$0.10 \\
~$\vdash$ w/o NLU & 12.85$\pm$0.18 & 40.65$\pm$0.37 & 18.24$\pm$0.26 & 8.13$\pm$0.10 & 30.80$\pm$0.36 & 13.10$\pm$0.09 \\
\bottomrule
\end{tabular}}}}
\label{tab:model-comparison}
\vspace{-0.1in}
\end{center}
\end{table*}

\begin{table}[t]
\begin{center}
\caption{Human Evaluations on \textsc{Gigaword}: \textit{InjType} (ours) v.s. Seq2Seq attention with type concatenation. }
\vspace{-0.1in}
\scalebox{0.90}{\begin{tabular}{l||ccc}
\toprule
 & Win & Tie & Lose \\
\midrule
Grammar & \textbf{25.2\%} & 50.4\% & 24.4\% \\
Fluency & \textbf{24.8\%} & 52.6\% & 22.6\% \\
Coherence & \textbf{48.0\%} & 21.4\% & 30.6\% \\
Informativeness & \textbf{50.6\%} & 21.2\% & 28.2\% \\
\bottomrule
\end{tabular}}
\label{tab:human-eval}
\vspace{-0.2in}
\end{center}
\end{table}

\subsection{Experimental Settings}
\noindent\textbf{Datasets.} We conduct experiments on two news datasets: \textsc{Gigawords} \cite{graff2003english}, \textsc{NYT} \cite{nyt}.
Statistics can be found in Table \ref{tab:datasets} in Appendix. To obtain entity types, we first tokenize the article then apply the Stanford NER extraction method in CoreNLP~\cite{finkel2005incorporating}. In total, there are 14 entity types: \textsc{Country}, \textsc{Location}, \textsc{Person}, \textsc{Weekday}, \textsc{Year}, \textsc{Month}, \textsc{Day}, \textsc{Organization}, \textsc{Timeunit}, \textsc{Digit}, \textsc{Digitrank}, \textsc{Digitunit}, \textsc{Timeunit}, \textsc{Lengthunit}.

\vspace{0.05in}
\noindent\textbf{Baselines.} We compare with three Seq2Seq methods (i.e., Seq2Seq~\cite{sutskever2014sequence}, SeqAttn~\cite{bahdanau2015neural}, CopyNet~\cite{gu2016incorporating}), 
and two pre-trained language models (i.e., GPT-2~\cite{radford2019language}, UniLM~\cite{dong2019unified}).
\textbf{It should be noted that} all Seq2Seq baselines are implemented with the concatenation of entity mention and entity types.
We fine-tune GPT-2 and UniLM on the training set for 20 epochs. Since our model is not pre-trained on large corpora and has much less model parameters, competing with GPT-2 and UniLM is very challenging. 

\vspace{0.05in}
\noindent\textbf{Implementation Details}
We take 256 as dimension size of hidden states for GRU encoder and decoder. The word embedding size is 300. We use Adam optimizer~\cite{kingma2014adam} with learning rate of 1e-4. We trained our model for 60 epochs on an NVIDIA 2080-Ti GPU. We did grid search on the hyperparameters in loss Eq.(\ref{eq:loss}) and got the best performance at $\lambda_1 = \lambda_2 = 2$.

\vspace{0.05in}
\noindent\textbf{Evaluation metrics.}
The performance is measured by standard corpus-level metrics, including BLEU~\cite{papineni2002bleu}, ROUGE~\cite{lin2004rouge}.

\subsection{Experimental Results}

Tables~\ref{tab:model-comparison} compares our \textit{InjType} with competitive baseline methods on three datasets. As shown in the table, \textit{InjType} performs the best among all methods.
This demonstrates that our proposed multi-step decoder is a more effective strategy than simple entity type embedding concatenation.

Table \ref{tab:case} shows a case in the test set to compare the generated results from different models. We observe our model can generate \emph{more precise contextual words} than baseline methods. 

We compare our model with its variants in Table~\ref{tab:model-comparison}. We observe that the mention predictor (MP) contributes more than NLU modules.
Adding MP improves BLEU-4 by +1.65\%, while adding NLU improves BLEU-4 by +0.26\%. The combination the two performs the best. The NLU module has a positive impact but not being claimed as a core contribution. It reuses the decoder to predict entity mentions and aligns seamlessly with our goal of effectively embedding entity meaning into hidden states. The bi-directional GRU used in the NLU brings extra coherency into the model since it considers entities and contexts after the current token.

\begin{table*}[h]
\begin{center}
\caption{Case study. Our proposed \textit{InjType} can generate more precise contextual words than baseline methods.}
\vspace{-0.1in}
\scalebox{0.85}{
{\begin{tabular}{p{17.5cm}}
\toprule
\textbf{Input entities:} China ({Country}, {C}), Russia ({C}), WTO ({Organization}, {O}), Chinese ({C}), Wen\_Jiabao ({Person}, {P}), Friday ({Weekday}, {W}), Moscow ({L}), Interfax\_news\_agency ({O}) \\
\midrule
\textbf{Ground truth:} {China} \textbf{\textgreen{has agreed to}} back {Russia}'s entry into the {WTO}, {Chinese} prime minister {Wen Jiabao} said {Friday} after talks in {Moscow}, the {Interfax news agency} \textbf{\textgreen{reported}} . \\
\midrule
\textbf{Seq2Attn:} {China} has lodged a veto global \textred{WTO} plan to \textred{WTO} the global trade body at its senior {Chinese} counterpart {Wen Jiabao} on {Friday}, {Interfax news agency} reported, as saying from {Moscow}, \textred{Interfax news agency said, quoted by the Interfax news agency}. \textblue{[The second ``WTO'' should be ``Russia''. ``Interfax'' appeared multiple times.]} \\
\midrule
\textbf{CopyNet:} {China} and {Russia} want to \textred{sell {WTO} trade sanctions against the {World\_Trade\_Organisation}} , warning the proposed sanctions against the \textred{WTO agreement are at the WTO agreement we are concerned} , the report said after meeting with his meeting counterpart Wen\_Jiabao on {Friday} in {Moscow} , the {Interfax\_news\_agency} reported . \\
\midrule
\textbf{\textit{InjType} (Our method):} {China} \textbf{\textgreen{has agreed}} {Russia}'s bid to join the {WTO} opening, {Chinese} \textred{foreign} minister {Wen Jiabao} said {Friday} {Moscow}, \textbf{\textgreen{report quoted the state official from}} {Interfax news agency}. \\
\bottomrule
\end{tabular}}}
\label{tab:case}
\end{center}
\vspace{-0.15in}
\end{table*}

\subsection{Human Evaluation}

We sample 50 examples from \textsc{Gigaword} test. Every generated news is presented to 5 annotators on Amazon Mechanical Trunk (AMT). Annotators were presented with two news articles and asked to decide which one was better and which one was worse in order of grammar and fluency, coherence, and informativeness. The result is “win”, “lose” or “tie”. 
Table \ref{tab:human-eval} demonstrates the human evaluation results. \textit{InjType} can significantly outperform Seq2Seq attention with type concatenation on coherence and informativeness. So, entity type injection is a more effective way to leverage entity type information than a simple concatenation way.

%% file: 5conclusions.tex
Entity plays the key role of making the text coherent in news generation. In order to enhance the role of entity, we propose a novel multi-step decoder that can effectively embed the entity's meaning into decoding hidden states, making the generated words precise. Experiments on two news datasets demonstrate that our method can perform better than conventional type embedding concatenation.

%% file: 6appendix.tex
\section{Appendix}

\begin{table}[h]
\caption{Statistics of two datasets.}
\vspace{-0.15in}
\begin{center}
\scalebox{0.90}{\begin{tabular}{l|c|c}
\toprule
& \textsc{Gigawords} & \textsc{NYT} \\
\midrule
\#Articles & 7,903 & 8,371 \\
\# Training & 5,903 & 6,371 \\
\# Validation & 1,000 & 1,000 \\
\# Test & 1,000 & 1,000 \\
\#Mentions $|\mathcal{M}|$ & 14,645 & 15,300\\
\#Types $|\mathcal{T}|$ & 14 & 14 \\
\#Words $|\mathcal{V}|$ & 30,121 & 38,802 \\
Len. Input/Output & 14.0 / 86.1 & 10.6 / 78.6 \\
Output Sentences & 4.1 & 3.8 \\
\bottomrule
\end{tabular}}
\label{tab:datasets}
\end{center}
\end{table}

\subsection{Additional Dataset Information}
\label{sec:data}

We create two datasets from public sources: \textsc{Gigawords} \cite{graff2003english} and \textsc{NYT} \cite{nyt}.
The Gigawords dataset contains around 4 million human-written news articles from various famous news publishers such as the New York Times and the Washington Posts from 1994 to 2010. The NYT dataset contains news articles written and published by New York Times from January, 1987 to June, 2007. It also contains 650,000 article summaries. 
For both datasets, we use 1,000 articles for development, 1,000 for test, and the remaining for training. On average, the models are expected to predict more than 70 contextual words \emph{precisely} from a list of about 10 entities. 

To obtain entity types, we first tokenize the article then apply the Stanford NER extraction method in CoreNLP~\cite{finkel2005incorporating}. In total, there are 14 entity types: \textsc{Country}, \textsc{Location}, \textsc{Person}, \textsc{Weekday}, \textsc{Year}, \textsc{Month}, \textsc{Day}, \textsc{Organization}, \textsc{Timeunit}, \textsc{Digit}, \textsc{Digitrank}, \textsc{Digitunit}, \textsc{Timeunit}, \textsc{Lengthunit}.

\begin{figure*}[h]
\centering
{\includegraphics[width=1.0\textwidth]{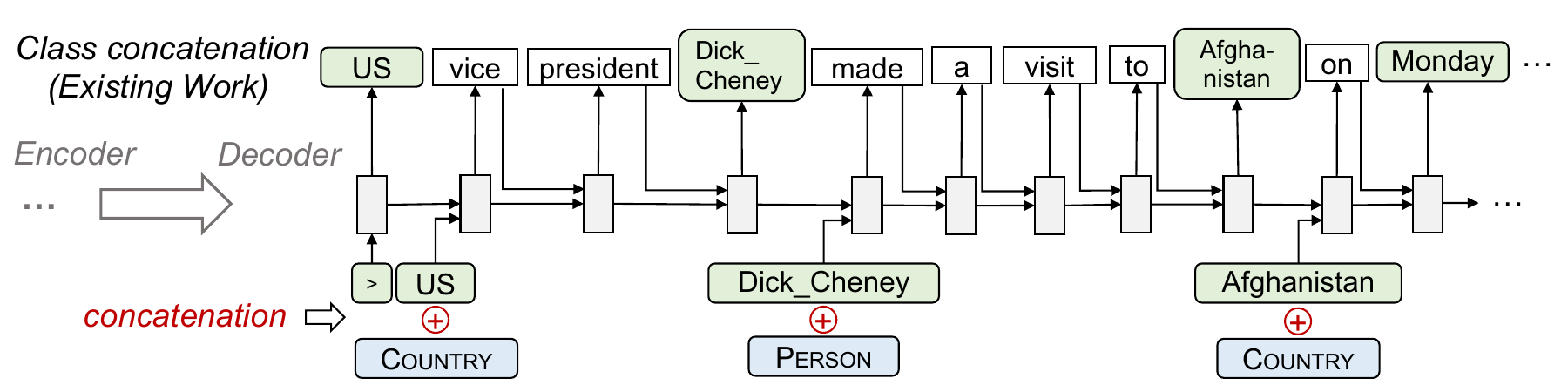}
\vspace{-0.3in}
\caption{Concatenating entity mention embeddings and type embeddings is a straightforward strategy to use the type information. However, it may not be effective due to lack of contextual information in the hidden states. \textbf{\textit{Note}} that the figure only highlights the decoder. Attention mechanism is also employed, but is not shown in the figure.}
\label{fig:framework_concatenation}}
\end{figure*}

\subsection{Entity Type Embedding Concatenation}

A natural way to incorporate type information into a Seq2Seq generation framework is to concatenate entity mention embeddings and type embeddings~\cite{yu2018typesql,zhao2019personalized,chan2019stick}. In the encoding phase, we take both entity mention and entity type as input, and learn contextual representation of each entity in the given list through a bi-directional GRU encoder. In the decoding phase, we adopt a standard attention mechanism~\cite{bahdanau2015neural} to generate output sequence. Specifically, in the \textit{encoding} phase, we concatenate the embedding of entity mention $x^M_i$ with the embeddings of its corresponding type $x^T_i$ extracted by CoreNLP~\cite{finkel2005incorporating}. So the input embedding of the $i$-th entity is defined as:
\begin{equation}
    \textbf{x}_t = \textbf{x}^M_t \oplus \textbf{x}^T_t, \nonumber
\end{equation}
where $\oplus$ denotes vector concatenation. We adopt bi-directional gated recurrent unit (Bi-GRU) \cite{cho2014learning} as encoder to capture contextualized representation for each entity given in the input list. The encoder has a forward GRU which reads input entities $X$ from $x_1$ to $x_n$, where $n$ is the length of input list. It also has a backward GRU to learn from the backward direction. We then concatenate hidden states from both directions:
\begin{equation}
    \textbf{h}_i = \overrightarrow{\mathrm{GRU}}(\textbf{x}_i) \oplus \overleftarrow{\mathrm{GRU}}(\textbf{x}_i). \nonumber
\end{equation}

In the \textit{decoding} phase, another GRU serves as the model decoder to generate entity mention and contextual words to generate the target sequence $\textbf{s}_t = \overrightarrow{\mathrm{GRU}}(\textbf{y}_t, \textbf{s}_{t-1}, \textbf{c}_t)$,
where $\textbf{y}_t$ is the concatenation of entity mention embedding $\textbf{y}^M_i$ with the embedding of its corresponding type $\textbf{y}^T_i$ (as shown in Figure \ref{fig:framework_concatenation}). So, given the current decoding hidden state ${\textbf{s}_t}$ and the source-side attentive context vector ${\textbf{c}_t}$, the readout hidden state is defined as ${\textbf{r}_t} = \mathrm{tanh}(\textbf{W}_c \cdot [{\textbf{s}_t} \oplus {\textbf{c}_t}])$, where the source-side attentive context vector ${\textbf{c}_t}$ at the current decoding step $t$ is computed through attention mechanism which matches the last decoding state ${\textbf{s}_{t-1}}$ with each encoder hidden state ${\textbf{h}_i}$ to get an importance score $\alpha_{t,i}$. Then, all importance scores ${\textbf{e}_{t,i}}$ are then normalized to get the current context vector $\textbf{c}_t$ by weighted sum:
\begin{align}
    & {\textbf{e}_{t,i}}  = \mathrm{tanh}(\textbf{W}_a \cdot {\textbf{s}_{t-1}} + \textbf{U}_a \cdot {\textbf{h}_i}),  \nonumber \\
    {\textbf{c}_t}  = \sum_{i=1}^n & \alpha_{t,i} \cdot \textbf{h}_i , ~\text{where}~ \alpha_{t,i}  = \frac{\exp({\textbf{e}_{t,i}})}{\sum_{i=1}^n \exp({\textbf{e}_{t,i}})}, \nonumber
\end{align}
where $\textbf{W}_a$ and $\textbf{U}_a$ are trainable parameters. Then the readout state $\textbf{r}_t$, is passed through a multilayer perceptron (MLP) to predict the next word with a softmax layer over the decoder vocabulary (all entity mentions and contextual words):
\begin{equation}
p(y_t \vert y_{<t},X) = \mathrm{softmax}(\textbf{W}_r \cdot {\textbf{r}_t}). \nonumber
\end{equation}

The loss function is defined as:
\begin{equation}
    \mathcal{L} = - \sum^m_{t=1} \log \left( p(y_t \in \mathcal{M} \cup \mathcal{V}|y_{< t}, X) \right). \nonumber
\label{kbnll}
\end{equation}

Figure~\ref{fig:framework_concatenation} is a reference to the model design.

\begin{table*}[h]
\begin{center}
\caption{Case study. Our proposed \textit{InjType} can generate more precise contextual words than baseline methods.}
\vspace{-0.1in}
\scalebox{0.75}{
{\begin{tabular}{p{21cm}}
\toprule
\textbf{Input entities:} British ({Country}, {C}), Gordon\_Brown ({Person}, {P}), United\_States ({C}), US ({C}), George\_W.\_Bush ({P}), White\_House ({Organization}, {O}), Thursday ({Weekday}, {W}), Brown ({P}), Camp\_David ({Location}, {L}), Sunday ({W}), Bush ({P}), Tony\_Snow ({P}). \\
\midrule
\textbf{Ground truth:} \textblue{British} prime minister \textblue{Gordon\_Brown} is to \textbf{\textgreen{make his first official trip}} as premier to the \textblue{United\_States} , for two days of talks with \textblue{US} president \textblue{George\_W.\_Bush} , the \textblue{White\_House} said \textblue{Thursday} . \textblue{Brown} \textbf{\textgreen{was to arrive at}} \textblue{Camp\_David} late \textblue{Sunday} , have dinner with \textblue{Bush} , `` then there will be a pretty full meeting schedule the following day , '' spokesman \textblue{Tony\_Snow} said . \\
\midrule
\textbf{Seq2Attn:} \textblue{British} prime minister \textblue{Gordon\_Brown} has recalled the \textblue{United\_States} for the highest \textblue{United\_States} since two days of \textblue{US} president \textblue{George\_W.\_Bush} 's close to the \textblue{White\_House} , the \textblue{White\_House} said \textblue{Thursday} . \textblue{Brown} , who on his cabinet reshuffle , arrived at the \textblue{White\_House} later this week before returning to a dinner at \textblue{Sunday} 's session . \\
\midrule
\textbf{CopyNet:} \textblue{British} prime minister \textblue{Gordon\_Brown} returned to the the \textblue{United\_States} head of two days before he will meet with \textblue{US} president \textblue{George\_W.\_Bush} , the \textblue{White\_House} said \textblue{Thursday} . \textblue{Brown} won the grand whereabouts at the weekend inauguration and his opponent was taking the g8 summit in this year 's retreat \textblue{Sunday} , said he would make his inauguration . \\
\midrule
\textbf{\textit{InjType} (Our method):} \textblue{British} prime minister \textblue{Gordon\_Brown} will \textbf{\textgreen{make his first official visit}} to the \textblue{United\_States} two days after talks with new \textblue{US} president \textblue{George\_W.\_Bush} , the \textblue{White\_House} said \textblue{Thursday} . \textblue{Brown} \textbf{\textgreen{has arrived at}} \textblue{Camp\_David} on \textblue{Sunday} as spokesman \textblue{Bush} said in a brief statement . `` today , something how long , '' spokesman \textblue{Tony\_Snow} said . \\ 
\bottomrule
\end{tabular}}}
\label{tab:case-2}
\end{center}
\vspace{-0.12in}
\end{table*}

\subsection{Baseline Methods}

\vspace{0.02in}
\noindent\textbf{Seq2Seq}~\cite{sutskever2014sequence} It is the basic encoder-decoder model widely used in NLG tasks.
 
\vspace{0.02in}
\noindent\textbf{SeqAttn}~\cite{bahdanau2015neural} It adds a soft attention mechanism to 
the input sequence where the most relevant information is concentrated.

\vspace{0.02in}
\noindent\textbf{CopyNet}~\cite{gu2016incorporating}\textbf{.} It introduces copy mechanism to decoder to alleviate the out-of-vocabulary issue caused by infrequent words.

\vspace{0.02in}
\noindent\textbf{GPT-2} ~\cite{radford2019language} It is a pre-trained Transformer language model. It excels at generating convincing articles with initial prompts.

\vspace{0.02in}
\noindent\textbf{UniLM} ~\cite{dong2019unified} It is also a pre-trained language model. It unifies three tasks: Seq2Seq generation, conditional generation, and NLU.

\subsection{Evaluation Metrics}

We use four kinds of standard metrics: 
(1) BLEU-4 measures the average 4-gram precision on a set of reference texts; (2) ROUGE-2 computes the recall of bigrams; (3) ROUGE-L computes the overlap of the longest common subsequence between the hypothesis and the reference; 

\subsection{Human Evaluation Settings}

We sample 50 inputs from {Gigaword} test set, and generate news articles by type embedding concatenation and our proposed \textit{InjType}.
Annotators were presented with two news articles and asked to decide which one was better and which one was worse in order of grammar and fluency (is the news written in well-formed English?), coherence (does the news describe the whole event coherently?), and informativeness (does the news relate to the input entities?). The result is “win”, “lose” or “tie”. 
We provided good and bad examples and explain how they succeed or fail to meet the defined criteria.
Every generated news is presented to 5 annotators on Amazon Mechanical Trunk (AMT).

%% file: naacl2021.bbl
\begin{thebibliography}{32}
\expandafter\ifx\csname natexlab\endcsname\relax\def\natexlab#1{#1}\fi

\bibitem[{Amplayo et~al.(2018)Amplayo, Lim, and Hwang}]{amplayo2018entity}
Reinald~Kim Amplayo, Seonjae Lim, and Seung-won Hwang. 2018.
\newblock Entity commonsense representation for neural abstractive
  summarization.
\newblock In \emph{North American Chapter of the Association for Computational
  Linguistics (NAACL)}.

\bibitem[{Bahdanau et~al.(2015)Bahdanau, Cho, and Bengio}]{bahdanau2015neural}
Dzmitry Bahdanau, Kyunghyun Cho, and Yoshua Bengio. 2015.
\newblock Neural machine translation by jointly learning to align and
  translate.
\newblock \emph{International Conference for Learning Representation (ICLR)}.

\bibitem[{Chan et~al.(2019)Chan, Chen, Wang, Li, Zhang, Gai, Zhao, and
  Yan}]{chan2019stick}
Zhangming Chan, Xiuying Chen, Yongliang Wang, Juntao Li, Zhiqiang Zhang, Kun
  Gai, Dongyan Zhao, and Rui Yan. 2019.
\newblock Stick to the facts: Learning towards a fidelity-oriented e-commerce
  product description generation.
\newblock In \emph{Conference on Empirical Methods in Natural Language
  Processing and International Joint Conference on Natural Language Processing
  (EMNLP-IJCNLP)}.

\bibitem[{Cho et~al.(2014)Cho, van Merri{\"e}nboer, Gulcehre, Bahdanau,
  Bougares, Schwenk, and Bengio}]{cho2014learning}
Kyunghyun Cho, Bart van Merri{\"e}nboer, Caglar Gulcehre, Dzmitry Bahdanau,
  Fethi Bougares, Holger Schwenk, and Yoshua Bengio. 2014.
\newblock Learning phrase representations using rnn encoder--decoder for
  statistical machine translation.
\newblock In \emph{Conference on Empirical Methods in Natural Language
  Processing}.

\bibitem[{Clark et~al.(2018)Clark, Ji, and Smith}]{clark2018neural}
Elizabeth Clark, Yangfeng Ji, and Noah~A Smith. 2018.
\newblock Neural text generation in stories using entity representations as
  context.
\newblock In \emph{Proceedings of the 2018 Conference of the North American
  Chapter of the Association for Computational Linguistics: Human Language
  Technologies (NAACL-HLT)}.

\bibitem[{Dong et~al.(2019)Dong, Yang, Wang, Wei, Liu, Wang, Gao, Zhou, and
  Hon}]{dong2019unified}
Li~Dong, Nan Yang, Wenhui Wang, Furu Wei, Xiaodong Liu, Yu~Wang, Jianfeng Gao,
  Ming Zhou, and Hsiao-Wuen Hon. 2019.
\newblock Unified language model pre-training for natural language
  understanding and generation.
\newblock In \emph{Advances in Neural Information Processing Systems
  (Neruips)}.

\bibitem[{Feng et~al.(2018)Feng, Liu, Liu, Qin, Sun, and Liu}]{feng2018topic}
Xiaocheng Feng, Ming Liu, Jiahao Liu, Bing Qin, Yibo Sun, and Ting Liu. 2018.
\newblock Topic-to-essay generation with neural networks.
\newblock In \emph{International Joint Conference on Artificial Intelligence
  (IJCAI)}.

\bibitem[{Finkel et~al.(2005)Finkel, Grenager, and
  Manning}]{finkel2005incorporating}
Jenny~Rose Finkel, Trond Grenager, and Christopher Manning. 2005.
\newblock Incorporating non-local information into information extraction
  systems by gibbs sampling.
\newblock In \emph{Annual meeting on association for computational linguistics
  (ACL)}.

\bibitem[{Graff et~al.(2003)Graff, Kong, Chen, and Maeda}]{graff2003english}
David Graff, Junbo Kong, Ke~Chen, and Kazuaki Maeda. 2003.
\newblock English gigaword.
\newblock \emph{Linguistic Data Consortium, Philadelphia}.

\bibitem[{Grosz et~al.(1995)Grosz, Joshi, and Weinstein}]{grosz1995centering}
Barbara~J Grosz, Aravind Joshi, and Scott Weinstein. 1995.
\newblock A framework for modeling the local coherence of discourse.
\newblock \emph{Computational Linguistics}.

\bibitem[{Gu et~al.(2016)Gu, Lu, Li, and Li}]{gu2016incorporating}
Jiatao Gu, Zhengdong Lu, Hang Li, and Victor~OK Li. 2016.
\newblock Incorporating copying mechanism in sequence-to-sequence learning.
\newblock In \emph{Annual Meeting of the Association for Computational
  Linguistics}.

\bibitem[{Ji et~al.(2017)Ji, Tan, Martschat, Choi, and Smith}]{ji2017dynamic}
Yangfeng Ji, Chenhao Tan, Sebastian Martschat, Yejin Choi, and Noah~A Smith.
  2017.
\newblock Dynamic entity representations in neural language models.
\newblock In \emph{Conference on Empirical Methods in Natural Language
  Processing (EMNLP)}.

\bibitem[{Kingma and Ba(2015)}]{kingma2014adam}
Diederik~P Kingma and Jimmy Ba. 2015.
\newblock Adam: A method for stochastic optimization.
\newblock \emph{International Conference for Learning Representation (ICLR)}.

\bibitem[{Lin(2004)}]{lin2004rouge}
Chin-Yew Lin. 2004.
\newblock Rouge: A package for automatic evaluation of summaries.
\newblock In \emph{Text summarization branches out}.

\bibitem[{Nenkova(2008)}]{nenkova2008entity}
Ani Nenkova. 2008.
\newblock Entity-driven rewrite for multi-document summarization.
\newblock In \emph{International Joint Conference on Natural Language
  Processing}.

\bibitem[{Papineni et~al.(2002)Papineni, Roukos, Ward, and
  Zhu}]{papineni2002bleu}
Kishore Papineni, Salim Roukos, Todd Ward, and Wei-Jing Zhu. 2002.
\newblock Bleu: a method for automatic evaluation of machine translation.
\newblock In \emph{Annual meeting on association for computational linguistics
  (ACL)}.

\bibitem[{Puduppully et~al.(2019)Puduppully, Dong, and
  Lapata}]{puduppully2019data}
Ratish Puduppully, Li~Dong, and Mirella Lapata. 2019.
\newblock Data-to-text generation with entity modeling.
\newblock In \emph{Annual Meeting of the Association for Computational
  Linguistics (ACL)}.

\bibitem[{Qin et~al.(2019)Qin, Galley, Brockett, Liu, Gao, Dolan, Choi, and
  Gao}]{qin2019conversing}
Lianhui Qin, Michel Galley, Chris Brockett, Xiaodong Liu, Xiang Gao, William~B
  Dolan, Yejin Choi, and Jianfeng Gao. 2019.
\newblock Conversing by reading: Contentful neural conversation with on-demand
  machine reading.
\newblock In \emph{Annual Meeting of the Association for Computational
  Linguistics (ACL)}.

\bibitem[{Radford et~al.(2019)Radford, Wu, Child, Luan, Amodei, and
  Sutskever}]{radford2019language}
Alec Radford, Jeffrey Wu, Rewon Child, David Luan, Dario Amodei, and Ilya
  Sutskever. 2019.
\newblock Language models are unsupervised multitask learners.
\newblock \emph{OpenAI blog}.

\bibitem[{Sandhaus(2008)}]{nyt}
Evan Sandhaus. 2008.
\newblock The new york times annotated corpus.
\newblock \emph{Linguistic Data Consortium, Philadelphia}.

\bibitem[{Sharma et~al.(2019)Sharma, Huang, Hu, and Wang}]{sharma2019entity}
Eva Sharma, Luyang Huang, Zhe Hu, and Lu~Wang. 2019.
\newblock An entity-driven framework for abstractive summarization.
\newblock In \emph{Conference on Empirical Methods in Natural Language
  Processing and International Joint Conference on Natural Language Processing
  (EMNLP-IJCNLP)}.

\bibitem[{Sutskever et~al.(2014)Sutskever, Vinyals, and
  Le}]{sutskever2014sequence}
Ilya Sutskever, Oriol Vinyals, and Quoc~V Le. 2014.
\newblock Sequence to sequence learning with neural networks.
\newblock In \emph{Advances in neural information processing systems
  (Neruips)}.

\bibitem[{Wang and Wan(2018)}]{wang2018sentigan}
Ke~Wang and Xiaojun Wan. 2018.
\newblock Sentigan: generating sentimental texts via mixture adversarial
  networks.
\newblock In \emph{International Joint Conference on Artificial Intelligence
  (IJCAI)}.

\bibitem[{Yang et~al.(2019)Yang, Li, Luo, Liu, and Sun}]{yang2019enhancing}
Pengcheng Yang, Lei Li, Fuli Luo, Tianyu Liu, and Xu~Sun. 2019.
\newblock Enhancing topic-to-essay generation with external commonsense
  knowledge.
\newblock In \emph{Annual Meeting of the Association for Computational
  Linguistics (ACL)}.

\bibitem[{Yao et~al.(2019)Yao, Peng, Weischedel, Knight, Zhao, and
  Yan}]{yao2019plan}
Lili Yao, Nanyun Peng, Ralph Weischedel, Kevin Knight, Dongyan Zhao, and Rui
  Yan. 2019.
\newblock Plan-and-write: Towards better automatic storytelling.
\newblock In \emph{Proceedings of the AAAI Conference on Artificial
  Intelligence (AAAI)}.

\bibitem[{Yu et~al.(2018)Yu, Li, Zhang, Zhang, and Radev}]{yu2018typesql}
Tao Yu, Zifan Li, Zilin Zhang, Rui Zhang, and Dragomir Radev. 2018.
\newblock Typesql: Knowledge-based type-aware neural text-to-sql generation.
\newblock In \emph{Conference of the North American Chapter of the Association
  for Computational Linguistics: Human Language Technologies (NAACL-HLT)}.

\bibitem[{Yu et~al.(2020)Yu, Zhu, Li, Hu, Wang, Ji, and Jiang}]{yu2020survey}
Wenhao Yu, Chenguang Zhu, Zaitang Li, Zhiting Hu, Qingyun Wang, Heng Ji, and
  Meng Jiang. 2020.
\newblock A survey of knowledge-enhanced text generation.
\newblock \emph{arXiv preprint arXiv:2010.04389}.

\bibitem[{Yu et~al.(2021)Yu, Zhu, Zhao, Guo, and Jiang}]{yu2021sentence}
Wenhao Yu, Chenguang Zhu, Tong Zhao, Zhichun Guo, and Meng Jiang. 2021.
\newblock Sentence-permuted paragraph generation.
\newblock In \emph{Conference on Empirical Methods in Natural Language
  Processing (EMNLP)}.

\bibitem[{Zeng et~al.(2021)Zeng, Lin, Yu, Cleland-Huang, and
  Jiang}]{zeng2021enhancing}
Qingkai Zeng, Jinfeng Lin, Wenhao Yu, Jane Cleland-Huang, and Meng Jiang. 2021.
\newblock Enhancing taxonomy completion with concept generation via fusing
  relational representations.
\newblock In \emph{ACM SIGKDD International Conference on Knowledge Discovery
  \& Data Mining (KDD)}.

\bibitem[{Zhang et~al.(2020)Zhang, Wang, Li, Gan, Brockett, and
  Dolan}]{zhang2020pointer}
Yizhe Zhang, Guoyin Wang, Chunyuan Li, Zhe Gan, Chris Brockett, and Bill Dolan.
  2020.
\newblock Pointer: Constrained text generation via insertion-based generative
  pre-training.
\newblock \emph{In Empirical Methods in Natural Language Processing (EMNLP)}.

\bibitem[{Zhao et~al.(2019)Zhao, Fu, Song, Sakai, Chen, Xie, and
  Qian}]{zhao2019personalized}
Guoshuai Zhao, Hao Fu, Ruihua Song, Tetsuya Sakai, Zhongxia Chen, Xing Xie, and
  Xueming Qian. 2019.
\newblock Personalized reason generation for explainable song recommendation.
\newblock \emph{ACM Transactions on Intelligent Systems and Technology (TIST)}.

\bibitem[{Zheng et~al.(2017)Zheng, Wang, Chen, and
  Sangaiah}]{zheng2017automatic}
Hai-Tao Zheng, Wei Wang, Wang Chen, and Arun~Kumar Sangaiah. 2017.
\newblock Automatic generation of news comments based on gated attention neural
  networks.
\newblock \emph{IEEE Access}.

\end{thebibliography}
